\documentclass{llncs}
\usepackage{graphicx}
\usepackage{float}
\graphicspath{ {.} }
\begin{document}

\title{END-TO-END ATTENTION-BASED DISTANT SPEECH RECOGNITION\\WITH HIGHWAY LSTM}
\titlerunning{Attention-based DSR}  
%
\author{Hassan Taherian}
\authorrunning{Hassan Taherian} 
%
\tocauthor{Hassan Taherian}
\institute{Laboratory for Intelligent Multimedia Processing,\\
Amirkabir University of Technology, Hafez St., Tehran, Iran\\
\email{hasan.t@aut.ac.ir}
\\[12pt]
\textbf{Technical Report}\\
September 2016}

\maketitle              

\begin{abstract}
End-to-end attention-based models have been shown to be competitive alternatives to conventional DNN-HMM models in the Speech Recognition Systems. In this paper, we extend existing end-to-end attention-based models that can be applied for Distant Speech Recognition (DSR) task. Specifically, we propose an end-to-end attention-based speech recognizer with multichannel input that performs sequence prediction directly at the character level. To gain a better performance, we also incorporate Highway long short-term memory (HLSTM) which outperforms previous models on AMI distant speech recognition task.
\end{abstract}

\section{Introduction}
Recently end-to-end neural network speech recognition systems has shown promising results \cite{hann,grav06,grav14}. Further improvement were reported by using more advanced models such as Attention-based Recurrent Sequence Generator (ARSG) as part of an end-to-end speech recognition system \cite{bahd}.

Although these new techniques help to decrease the word error rate (WER) on the automatic speech recognition system (ASR), Distant Speech Recognition (DSR) remains a challenging task due to the reverberation and overlapping acoustic signals, even with sophisticated front-end processing techniques such as state-of-the-art beamforming \cite{kum}.

It is reported that using multiple microphones with signal preprocessing techniques like beamforming will improve the performance of the DSR \cite{kum}. However the performance of such techniques are suboptimal since they are depended heavily on the microphone array geometry and the location of target source \cite{selt}. Other works have shown that Deep Neural Networks with multichannel inputs can be used for learning a suitable representation for distant speech recognition without any front-end preprocessing \cite{arnab}. 

In this paper, we propose an extension to the current Attention-based Recurrent Sequence Generator (ARSG) that can handle multichannel inputs. We use \cite{bahd} as the baseline while extend its single channel input to multiple channel. We also integrate the language model in the decoder of ASRG with Weighted Finite State Transducer (WFST) framework as it was purposed in \cite{bahd}. To avoid slowness in convergence, we use Highway LSTM \cite{glass} in our model which allows us to have stacked LSTM layers without having gradient vanishing problem.

Rest of the paper is organized as follows. In Section 2 we briefly discuss related work. In Section 3 we introduce our proposed model and describe integration of language model with Attention-based Recurrent Sequence Generator. Finally, Conclusions are drawn in Section 4.

\section{Related Work}
Kim et al. already proposed an attention-based approach with multichannel input \cite{kim}. In their work an attention mechanism is embeded within a Recurrent Neural Network based acoustic model in order to weigh channel inputs for a number of frames. The more reliable channel input will get higher score. By calculating phase difference between microphone pairs, they also incorporated spatial information in their attention mechanism to accelerate learning of auditory attention. They reported that this model achieve comparable performance to beamforming techniques. 

One advantage of using such model is that no prior knowledge of microphone array geometry is required and real time processing is done much faster than models with front-end preprocessing techniques. However, by using conventional DNN-HMM acoustic model, they did not utilize the full potential of attention mechanism in their proposed model. The feature vectors, weighted by attention mechanism, can help to condition the generation of the next element of the output sequence. In our model, attention mechanism is similar to \cite{liu} but it I also connected to the ARGS to produce an end-to-end results.

\section{Proposed Model}
Our model consists of three parts: an encoder, an attention mechanism and a decoder. The model takes multichannel input X and stochastically generates an output sequence ($y_1$, $y_2$, ..., $y_T$). The input X consists of N channel inputs X=\{$X^{ch1}$, $X^{ch2}$,..., $X^{chN}$\} where each channel $X^{chi}$ is a sequence of small overlapping window of audio frames $X^{chi}$=\{$X_1^{chi}$, $X_2^{chi}$,..., $X_L^{chi}$\}. The encoder transforms the input into another representation by using Recurrent Neural Network (RNN). Then the attention mechanism weighs elements of new representation based on their relevance to the output $y$ at each time step. Finally, the decoder which is a RNN in our model, produces an output sequence ($y_1$, $y_2$, ..., $y_T$) one character at time by using weighted representation produced by attention mechanism and the hidden state of the decoder. In the literature, combination of the attention mechanism and the decoder is called Attention Recurrent Sequence Generator (ARSG). ARSG is proposed to address the problem of learning variable-length input and output sequences since in the conventional RNNs, the length of the sequence of hidden state vectors is always equal to the length of the input sequence.

In the following subsections we will discuss these three parts in detail.

\subsection{The Encoder}

We concatenate frames of all channels and feed them as an input to the encoder which is a RNN, at each time step t. Another approach to handle multichannel input as Kim et al. proposed in \cite{kim}, is  to feed each channel separately in to the RNN at each time step. Training such RNN in this approach can be very slow for an end-to-end scenario since the attention mechanism should weigh frames of each channel at every time step. Moreover, phase difference calculation between microphone pairs are needed for scoring frames. Because of these computational complexities, we decided to concatenate all channel features and feed them to the network as Liu et al. have done \cite{liu} in their model. 

\begin{figure}[h]
\centering
\includegraphics[scale=0.4]{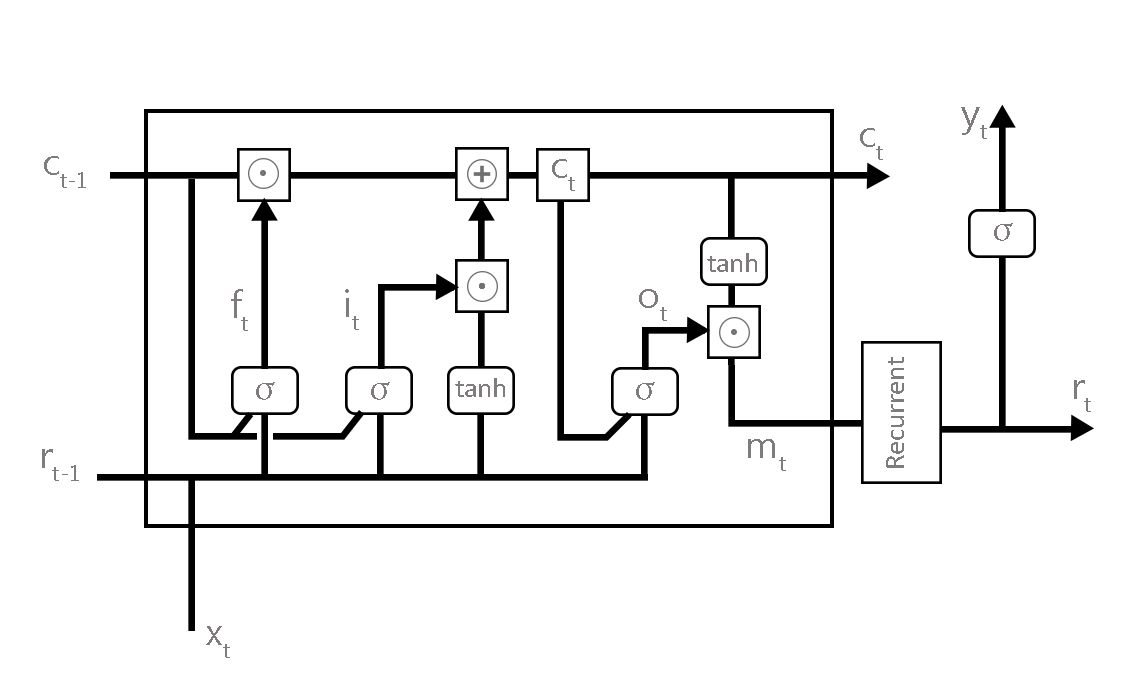}
\caption{The architecture of LSTM Projected (LSTMP). It is similar to LSTM with peephole with additioal projected layer. LSTMP reduces training parameters and allows to have deeper models}
\label{fig:lstmp}
\end{figure}

We used bidirectional highway long short-term memory (BHLSTM) in the encoder.  Zhang et al.  showed that BHLSTM with dropout achieved state-of-the-art performance in the AMI (SDM) task \cite {glass}. This model addresses  the gradient vanishing problem caused by stacking LSTM layers. Thus this model allows the network to go much deeper \cite {glass}. BHLSTM is based on long short-term memory projected (LSTMP) which originally proposed by \cite{gog}. The operation of LSTMP network follows the equations

\begin{eqnarray}
  i_t &=&  \sigma(\mathbf{W_{xi}}x_t + \mathbf{W_{mi}}r_{t-1} + \mathbf{W_{ci}}c_{t-1} + b_i )\\[4pt]
  f_t &=&  \sigma(\mathbf{W_{xf}}x_t + \mathbf{W_{mf}}r_{t-1} + \mathbf{W_{cf}}c_{t-1} + b_f )\\[4pt]
  c_t &=&  f_t \odot c_{t-1} + i_t \odot tanh(\mathbf{W_{xc}}x_t + \mathbf{W_{mc}}r_{t-1} + b_c ) \label{has:one} \\[4pt] 
  o_t &=&  \sigma(\mathbf{W_{xo}}x_t + \mathbf{W_{mo}}r_{t-1} + \mathbf{W_{co}}c_{t} + b_o )\\[4pt]
  m_t &=&  o_t \odot tanh(c_t)\\[4pt]
  r_t &=&  \mathbf{W_{rm}}m_t\\[4pt]
  y_t &=&  \sigma(\mathbf{W_{yr}}r_t + b_y) 
\end{eqnarray}

iteratively from $t=1$ to $T$,
where $\mathbf{W}$ term denote weight matrices. $\mathbf{W_{fc}}$, $\mathbf{W_{ic}}$, $\mathbf{W_{oc}}$ are diagonal weight matrices for peephole connections. the $b$ terms denote bias vectors, $\odot$ denotes the element-wise product, $i$, $f$, $o$, $c$ and $m$ are input gate, forget gate, output gate, cell activation and cell output activation vector respectively. $x_t$ and $y_t$ denotes input and output to the layer at time step t. $\sigma()$ is sigmoid activation fuction. finally $r$ denotes the recurrent unit activations. a LSTMP unit can be seen as combination of standard LSTM unit with peephole that its cell output activation $m$ is transformed by a linear projection layer. This architecture, shown in the Figure \ref{fig:lstmp}, converges faster than standard LSTM and it has less parameters than standard LSTM while keeping the performance \cite{gog}.\\

\begin{figure}[H]
\advance\leftskip-3cm
\includegraphics[scale=0.5]{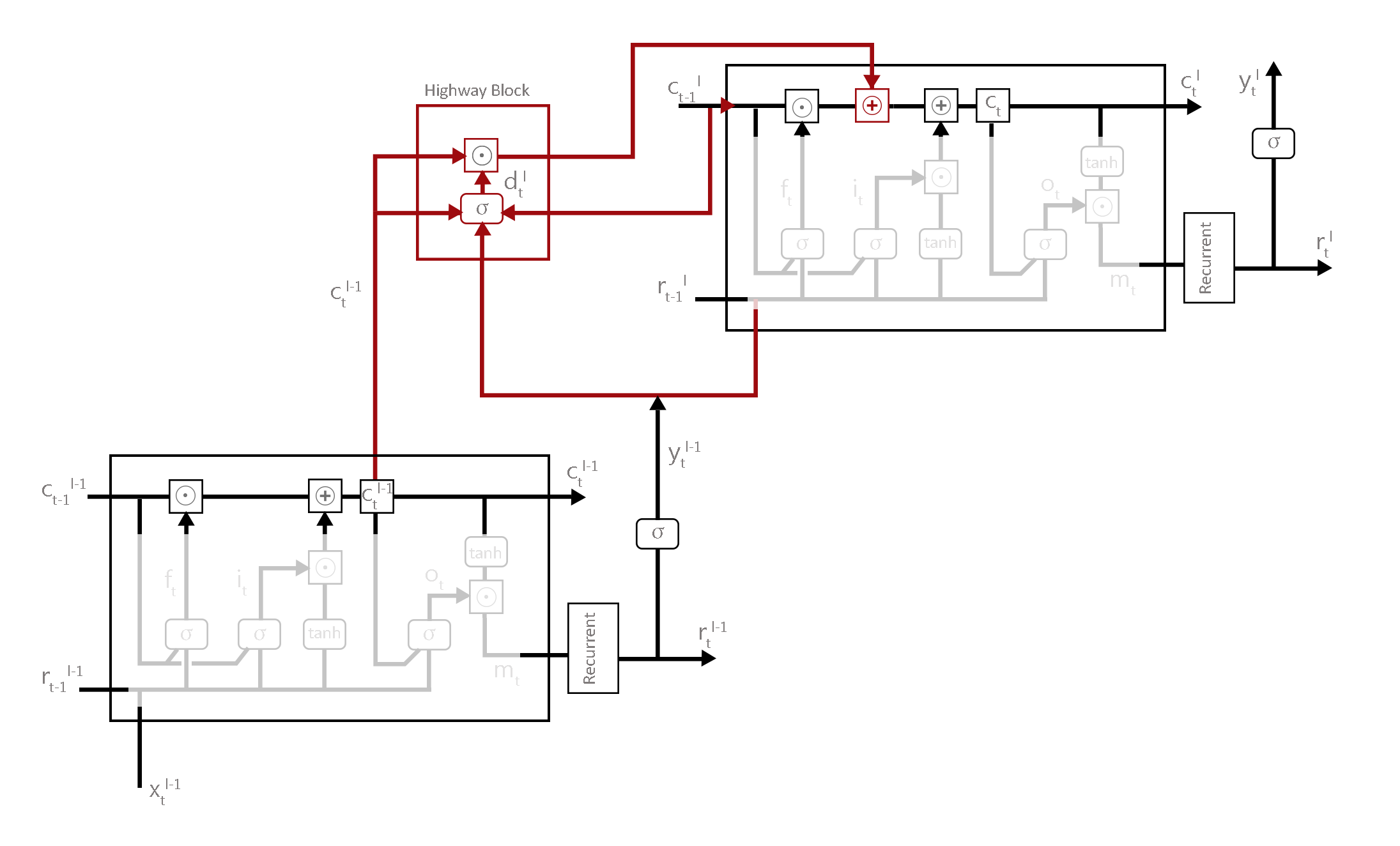}
\caption{The architecture of Highway LSTM (HLSTM). Cell states of adjacent layers with LSTMP nodes are connected with Highway Block. HLSTM has been shown to be effective to combat gradient vanishing problem in deep recurrent neural networks}
\label{fig:hlstm}
\end{figure}

The HLSTM RNN is illustrated in Figure \ref{fig:hlstm}. It is an extension to the LSTMP model. It has an extra gate, called carry gate, between memory cells of adjacent layers. The carry gate controls how much information can flow between cells in adjacent layers. HLSTM architecture can be orbtained by modification of Eq. (\ref{has:one}):

\begin{eqnarray}
  c_t^l &=&  f_t^l \odot c_{t-1}^l + i_t^l \odot tanh(\mathbf{W_{xc}^l}x_t^l + \mathbf{W_{mc}^l}r_{t-1}^l + b_c ) + d_t^l \odot c_t^{l-1}\\[4pt]
d_t^l &=&  \sigma(\mathbf{W_{xd}^l}x_t^l + w_{d1}^l \odot c_{t-1}^l + w_{d2}^l \odot c_{t}^{l-1} + b_d^l )
\end{eqnarray}

if the predecessor layer $(l−-1)$ is also an LSTM layer, otherwise:
\begin{eqnarray}
  c_t^l &=&  f_t^l \odot c_{t-1}^l + i_t^l \odot tanh(\mathbf{W_{xc}^l}x_t^l + \mathbf{W_{mc}^l}r_{t-1}^l + b_c ) + d_t^l \odot x_t^{l}\\[4pt]
d_t^l &=&  \sigma(\mathbf{W_{xd}^l}x_t^l + w_{d1}^l \odot c_{t-1}^l  + b_d^l )
\end{eqnarray}

where $b_d^l$ is a bias vector, $w_{d1}^l$ and $w_{d2}^l$ are weight vectors, $\mathbf{W_{xd}^l}$ is the weight matrix connecting the carry gate to the input of this layer. In this study, we also extend the HLSTM RNNs from unidirection to bidirection. Note that the backward layer follows the same equations used in the forward layer except that $t-1$ is replaced by $t+1$ to exploit future frames and the model operates from $t=T$ to 1. The output of the forward and backward layers are concatenated to form the input to the next layer.

\subsection{The Attetion Mechanism}

Attention mechanism is a subnetwork that weighs the encoded inputs $h$. It selects the temporal locations on input sequence that is useful for updating hidden state of the decoder. Specifically, at each time step $i$, the attention mechanism computes scalar energy $e_{i,l}$ for each time step of encoded input $h_l$ and hidden state of decoder $s_i$. The scalar energy will be converted into probability distribution, which is called alignment, over time steps using softmax function. Finally, the context vector $c_i$ is calculated by linear combination of elements of encoded inputs $h_l$ \cite{listen}.

\begin{eqnarray}
  e_{i,l} &=&  w^Ttanh(\mathbf{W}s_{i} + \mathbf{V}h_l + b)\label{eq:energy}\\[4pt]
  \alpha_{i,l} &=& \frac{exp(e_{i,l})}{\sum_{l=1}^L exp(e_i)}\\[4pt]
 c_{i} &=& \sum_{l} \alpha_{i,l}h_l
\end{eqnarray}

Where $\mathbf{W}$ and $\mathbf{V}$ are matrix parameters and $w$ and $b$ are vector parameters. Note that the $\alpha_{i,l}>0$ and  $\sum_{l} \alpha_{i}=1$. This method is called context-based. Bahdanau et al. showed that using context-based scheme for attention model is prone to error since similar elements of embedded input $h$ is scored equally \cite{bahd}. To alleviate this problem, Bahdanau et al. suggested to use previous alignment $\alpha_{i-1}$ by convolving it along time axis for calculation of the scalar energy. Thus by changing Eq. (\ref{eq:energy}) we have:

\begin{eqnarray}
  \mathbf{F} &=& \mathbf{Q}*\alpha_{i-1}\\[4pt]
  e_{i,l} &=&  w^Ttanh(\mathbf{W}s_{i} + \mathbf{V}h_l + \mathbf{U}f_l + b)
\end{eqnarray}

Where $\mathbf{U}$, $\mathbf{Q}$ are parameter matrices, * denotes convolution and $f_l$ are feature vectors of matrix $\mathbf{F}$. A disadvantage of this model is the complexity of the training procedure. The alignments $\alpha_{i,l}$ should be calculated for all pairs of input and output position which is –$\rm O(LT)$. Chorowski et al. suggested a windowing approach which limits the number of embedded inputs for computation of alignments \cite{chor}. This approach reduces the complexity to $\rm O(L+T)$.

\subsection{The Decoder}

The task of the decoder which is a RNN in ARSG framework, is to produce probability distribution over the next character conditioned on all the characters that already has been seen.  This distribution is generated with a MLP and a softmax by the hidden state of the decoder $s_i$ and the context vector of attention mechanism $c_i$.

\begin{eqnarray}
  \mathbf{P}(y_i|X,y_{<i}) &=& \mathbf{MLP}(s_i,c_i) \label{mlpdist}
\end{eqnarray}

The hidden state of decoder $s_i$ is a function of previous hidden state $s_{i-1}$, the previously emitted character $y_{i-1}$ and context vector of attention mechanism $c_{i-1}$:

\begin{eqnarray}
  s_i &=& f(s_{i-1},y_{i-1},c_{i-1})
\end{eqnarray}

Where $f$ is a single layer of standard LSTM. The ARSG with encoder can be trained jointly for end-to-end speech recognition. The objective function is to maximize the log probability of each character sequence condition on the previous characters.

\begin{eqnarray}
  max_{\theta} \sum_{i} log\ \mathbf{P}(y_i|X,y_{<i}^*;\theta)
\end{eqnarray}

Where $\theta$ is the network parameter and $y^*$ is the ground truth of previous characters. To make it more resilient to the bad predictions during test, Chen et al. suggested to sample from previous predicted characters with the rate of 10\% and use them as input for predicting next character instead of ground truth transcript \cite{listen}.

Chen et al. showed that end-to-end models for speech recognition can achieve good results without using language model but for reducing word error rate (WER), language model is essential \cite{listen}. One way for integrating language model with the ARSG framework, as Bahdanau et al. suggested, is to use Weighted Finite State Transducer (WFST) framework \cite{bahd}. By composition of grammar and lexicon, WFST defines the log probability for a character sequence ( see \cite{mohri}) . With the combination of WFST and ARGS, we can look for transcripts that maximize:

\begin{eqnarray}
  L &=& \sum_{i} log\ \mathbf{P}(y_i|X,y_{<i}^*;\theta) +\beta log_{P_{LM}}(y) 
\end{eqnarray}

in decoding time. Where $\beta$ is tunable parameter. (see \cite{bahd}). There are also other models for integrating language model to ARGS. Shallow Fusion and Deep Fusion are two of them that are proposed by Gulcehre et al. \cite{gulc}.

Integrating Shallow Fusion is similar to the WFST framework but it uses Recurrent Neural Network language model (RNNLM) for producing log probability of sequences. In Deep Fusion, the RNNLM is integrated with the decoder subnetwork in ARSG framework. Specifically, the hidden state of RNNLM is used as input for generating the probability distribution on the next character in Eq. (\ref{mlpdist}). However, in our work we opt to use WFST framework for the sake of simplicity and training tractability of the model.

\section{Conclusion}
In this work, we show the end-to-end models that are based only on neural networks can be applied to the Distant Speech Recognition task with concatenation of multichannel inputs. Also better accuracy is expected with the usage of Highway LSTMs in the encoder network. For the future work, we will study integration of Recurrent Neural Network Language Model (RNNLM) with Attention-based Recurrent Sequence Generator (ARSG) with Highway LSTM.

\end{document}